\documentclass[11pt,a4paper]{article}
\pdfoutput=1
\usepackage[nohyperref]{acl2017}  
\usepackage{times}
\usepackage{latexsym}

\usepackage{url}

\usepackage{amsmath}
\usepackage[margin=1in]{geometry}
\usepackage{amssymb}
\usepackage{multirow}
\usepackage{float}
\usepackage[]{algorithm2e}
\usepackage{array}
\usepackage{booktabs}
\usepackage{caption}
\usepackage{subcaption}
\usepackage{fixltx2e}
\usepackage{arydshln}

\usepackage{tikz}
\usetikzlibrary{shapes,arrows}

\newcommand{\ul}[1] {\underline{#1}}

\newcommand{\rel}[1] {{\it #1}}
\newcommand{\CosAdd}[0] {\textsc{3CosAdd}}
\newcommand{\PairDist}[0] {\textsc{PairwiseDistance}}
\newcommand{\CosMul}[0] {\textsc{3CosMul}}
\newcommand{\CosAvg}[0] {\textsc{3CosAvg}}

\newcommand{\LOne}[0] {form-of}
\newcommand{\LTwo}[0] {has-lab-number}
\newcommand{\LThree}[0] {has-tradename}
\newcommand{\LFour}[0] {tradename-of}
\newcommand{\LFive}[0] {associated-substance}
\newcommand{\LSix}[0] {has-free-acid-or-base-form}
\newcommand{\LSeven}[0] {has-salt-form}
\newcommand{\LEight}[0] {measured-component-of}
\newcommand{\HOne}[0] {refers-to}
\newcommand{\HTwo}[0] {same-type}
\newcommand{\MOne}[0] {adjectival-form-of}
\newcommand{\MTwo}[0] {noun-form-of}
\newcommand{\COne}[0] {associated-with-malfunction-of-gene-product}
\newcommand{\CTwo}[0] {gene-product-malfunction-associated-with-disease}
\newcommand{\CThree}[0] {causative-agent-of}
\newcommand{\CFour}[0] {has-causative-agent}
\newcommand{\CFive}[0] {has-finding-site}
\newcommand{\CSix}[0] {associated-with}
\newcommand{\AOne}[0] {anatomic-structure-is-part-of}
\newcommand{\ATwo}[0] {anatomic-structure-has-part}
\newcommand{\AThree}[0] {is-located-in}
\newcommand{\BOne}[0] {regulated-by}
\newcommand{\BTwo}[0] {regulates}
\newcommand{\BThree}[0] {gene-encodes-product}
\newcommand{\BFour}[0] {gene-product-encoded-by}

\newcommand{\relref}[2] {\rel{#1} (#2)}

\newcommand{\BMASS}[0] {BMASS}

\newcommand{\RelAcc}[0] {Acc$_R$}

\newcolumntype{P}[1]{>{\centering\arraybackslash}p{#1}}  
\newcolumntype{M}[1]{>{\centering\arraybackslash}m{#1}}  

\makeatletter
\def\hlinewd#1{
\noalign{\ifnum0=`}\fi\hrule \@height #1
\futurelet\reserved@a\@xhline}
\makeatother

\newlength{\Oldarrayrulewidth}

\aclfinalcopy 
\def\aclpaperid{19} 

\title{Insights into Analogy Completion from the Biomedical Domain}

\author{Denis Newman-Griffis$^{\dagger,\ddagger}$ \and
  Albert M. Lai$^{\clubsuit}$\and
  Eric Fosler-Lussier$^{\dagger}$ \\
  $^{\dagger}$The Ohio State University, Columbus, OH\\
  $^{\ddagger}$National Institutes of Health, Clinical Center, Bethesda, MD \\
  $^{\clubsuit}$Washington University in St. Louis, St Louis, MO\\
  {\tt newman-griffis.1@osu.edu, amlai@wustl.edu,}\\
  {\tt fosler@cse.ohio-state.edu }\\
}

\date{}

\begin{document}
\maketitle
\begin{abstract}
Analogy completion has been a popular task in recent years for evaluating
the semantic properties of word embeddings, but the standard methodology
makes a number of assumptions about analogies that do not always hold, 
either in recent benchmark datasets or when expanding into other domains.
Through an analysis of analogies in the
biomedical domain, we identify three assumptions: that of a {\it Single
Answer} for any given analogy, that the pairs involved describe the {\it Same Relationship},
and that each pair is {\it Informative} with respect to the other. We propose modifying
the standard methodology to relax these assumptions by allowing for multiple
correct answers, reporting MAP and MRR in addition to accuracy, and using multiple
example pairs.  We further present
BMASS, a novel dataset for evaluating linguistic regularities in biomedical
embeddings, and demonstrate that the relationships described in the dataset pose
significant semantic challenges to current word embedding methods.

\end{abstract}

\section{Introduction}

Analogical reasoning has long been a staple of computational semantics research, as it allows for evaluating how well
implicit semantic relations between pairs of terms are represented in a semantic model.  In particular, the recent boom
of research on learning vector space models (VSMs) for text \cite{Turney2010} has leveraged analogy completion as a standalone
method for evaluating VSMs without using a full NLP system.  This is due largely to the observations of ``linguistic regularities''
as linear offsets in context-based semantic models \cite{Mikolov2013c,Levy2014,Pennington2014}.

In the analogy completion task, a system is presented with an example term pair and a query, e.g.,
{\it London:England::Paris:\ul{\phantom{France}}}, and the task is to correctly fill in the blank.  Recent methods
consider the vector difference between related terms as representative of the relationship between them, and use this to
find the closest vocabulary term for a target analogy, e.g., {\it England - London + Paris $\approx$ France}.  However,
recent analyses reveal weaknesses of such offset-based methods, including that the use of cosine similarity often reduces to just
reflecting nearest neighbor structure \cite{Linzen2016}, and that there is significant variance in performance between different
kinds of relations \cite{Koper2015,Gladkova2016,Drozd2016}.

We identify three key assumptions encoded in the standard offset-based methodology for analogy completion: that a given
analogy has only
one correct answer, that all relationships between the example pair and the query-target pair are the same, and that
the example pair is sufficiently informative with respect to the query-target pair.  We demonstrate that these assumptions are
violated in real-world data, including in existing analogy datasets.  We then propose several modifications to
the standard methodology to relax these assumptions, including allowing for multiple correct answers, making use of multiple
examples when available, and reporting mean average precision (MAP) and mean reciprocal rank (MRR) to give a more complete
picture of the implicit ranking used in finding the best candidate for completing a given analogy.

Furthermore, we present the BioMedical Analogic Similarity Set (BMASS), a novel dataset for analogical reasoning in the
biomedical domain.  This new resource presents real-world examples of semantic relations of interest for biomedical
natural language processing research, and we hope it will support further research into biomedical VSMs \cite{Chiu2016b,
Choi2016KDD}.\footnote{
    The dataset, and all code used for our experiments, is available online at
    {\tt https://github.com/OSU-slatelab/BMASS}.
}

\section{Related work}

Analogical reasoning has been studied both on its own and as a component of downstream tasks, using a range
of systems.  Early work used rule-based systems for world knowledge \cite{Reitman1965} and syntactic \cite{Federici1997a}
relationships.  Supervised models were used for SAT (Scholastic Aptitude Test) analogies \cite{Veale2004}, and later for
synonymy, antonymy, and some world knowledge \cite{Turney2008,Herdagdelen2009}.
Analogical reasoning has also been used in support of downstream tasks, including word sense disambiguation \cite{Federici1997b}
and morphological analysis \cite{Lepage2009,Lavallee2010,Soricut2015}.

Recent work on analogies has largely focused on their use as an intrinsic evaluation of the properties of a VSM.
The analogy dataset of \newcite{Mikolov2013a}, often referred to as the Google dataset, has become a standard evaluation for
general-domain word embedding models \cite{Pennington2014,Levy2014,Schnabel2015,Faruqui2015}, and includes both world knowledge
and morphosyntactic relations.  Other datasets include the MSR analogies \cite{Mikolov2013c}, which describe morphological
relations only; and BATS \cite{Gladkova2016}, which includes both morphological and semantic relations.  The semantic relations from
SemEval-2012 Task 2 \cite{Jurgens2012} have also been used to derive analogies; however, as with the lexical Sem-Para dataset of
\newcite{Koper2015}, the semantic relationships tend to be significantly more challenging for embedding-based methods
\cite{Drozd2016}.  Additionally, \newcite{Levy2015a} demonstrate that even for some lexical relations where embeddings appear
to perform well, they are actually learning prototypicality as opposed to relatedness.

\section{Analogy completion task}

\subsection{Standard methodology}
\label{ssec:standard_methodology}

Given an analogy {\it a:b::c:d}, the evaluation task is to guess $d$ out of the vocabulary, given $a,b,c$ as evidence.  Recent
methods for this involve using the vector difference between embedded representations of the related pairs to rank all terms
in the vocabulary by how well they complete the analogy, and choosing the best fit.  The vector difference is most commonly
used in one of three ways, where $cos$ is cosine similarity:
\begin{align}
    \label{eq:3CosAdd}
    argmax_{d\in V}\hspace{0.1cm}&\big(cos(d, b-a+c)\big)\\
    \vspace{0.3cm}
    \label{eq:PairDist}
    argmax_{d\in V}\hspace{0.1cm}&\big(cos(d-c, b-a)\big)\\
    \label{eq:3CosMul}
    argmax_{d\in V}\hspace{0.1cm}&\frac{cos(d,b)cos(d,c)}{cos(d,a) + \epsilon}
\end{align}
Following the terminology of \newcite{Levy2014}, we refer to Equation~\ref{eq:3CosAdd} as \CosAdd, Equation~\ref{eq:PairDist} as
\PairDist, and Equation~\ref{eq:3CosMul} (which is equivalent to \CosAdd\ with log cosine similarities) as \CosMul.

In order to generate analogy data for this task, recent datasets have followed a similar process
\cite{Mikolov2013a,Mikolov2013c,Koper2015,Gladkova2016}.  First, relations of interest
were manually selected for the target domains: syntactic/morphological,
lexical (e.g., hypernymy, synonymy), or semantic (e.g., \rel{CapitalOf}).  Then, for each relation, example word pairs were manually
selected or automatically generated from existing resources (e.g., WordNet).  The final analogies were then generated by exhaustively
combining the sets of word pairs within each relation.

\subsection{Assumptions}
\label{ssec:assumptions}

Several key assumptions are inherent in this standard methodology that are not reflected in recent benchmark analogy datasets.
The first we refer to as the {\it Single-Target}
assumption: namely, that there is a single correct answer
for any given analogy.  Since the target $d$ is chosen via argmax, if we consider the following two analogies:
\begin{center}
    \it flu:nausea::fever:?cough\\
    flu:nausea::fever:?light-headedness
\end{center}
we must necessarily get at least one answer wrong.  \newcite{Gladkova2016} convert these analogies into a single case:
\begin{center}
    \it flu:nausea::fever:?[cough, light-headedness]
\end{center}
where either $cough$ or $light\-headedness$ is a correct guess.  However, this still misses our desire to get both correct
answers, if possible.  Relations with multiple correct targets are present in all of Google, BATS, and Sem-Para.

The second key assumption is that all the information relating $a$ to $b$ also relates $c$ to $d$.  While the pairs are
chosen based on a single common relationship, each pair may actually pertain to multiple relationships.  An example from
the Google dataset is {\it brother:sister::husband:wife}; Table~\ref{tbl:multi_relation} shows the semantic relations
involved in this analogy.  While the target relation \rel{FemaleCounterpart} is present in both pairs, by comparing the offsets
$sister-brother$ and $wife-husband$, we assume that either all ways in which each pair is related are present in both,
or that \rel{FemaleCounterpart} dominates the offset.  We refer to this as the {\it Same-Relationship} assumption.

\begin{table}[t]
    \centering
    \begin{tabular}{c|c}
        Pair&Relations\\
        \hline
        \multirow{2}{*}{brother:sister}&\rel{\bf FemaleCounterpart}\\
        &\rel{SiblingOf}\\
        \hline
        \multirow{2}{*}{husband:wife}&\rel{\bf FemaleCounterpart}\\
        &\rel{MarriedTo}
    \end{tabular}
    \caption{Binary semantic relations in ``brother is to sister as husband is to wife.''
        The target common relation is shown in bold.}
    \label{tbl:multi_relation}
\end{table}

Finally, it is not sufficient for two pairs to share a common relationship label; that relationship must be both representative
and informative for analogies to make sense (the {\it Informativity} assumption).
Relation labels may be sufficiently broad as to be meaningless, as we encountered when drawing unfiltered binary relations from
the Unified Medical Language System (UMLS) Metathesaurus.
One sample analogy from the \rel{RO:Null} relation (indicating ``related in some way'') was
{\it socks:stockings::Finns:Finnish language}.  While both pairs are of related terms, they are in no way related to
one another.

Furthermore, even when two pairs are examples of the same kind of clearly-defined relation, they may still be relatively
uninformative.  For example, in the Sem-Para \rel{Meronym} analogy {\it apricot:stone::trumpet:mouthpiece} the meronymic
relationship between $apricot$ and $stone$ could plausibly identify a number of parts of a trumpet: $mouthpiece$, $valves$,
$slide$, etc.\footnote{
    While this is similar to the Single-Target assumption, it bears separate consideration in that
    Single-Target refers to multiple valid objects of a specific relationship, while this is an issue of multiple valid
    relationships being described.
}  The extremely 
high-level nature of several of the Sem-Para relations (hypernymy, antonymy, and
synonymy) suggests that some of the difficulty observed by \newcite{Koper2015} is due to violations of Informativity.
\section{BMASS}
\label{sec:bmass}

We present \BMASS\ (the BioMedical Analogic Similarity Set), a dataset of biomedical analogies, generated using the expert-curated
knowledge in the Unified Medical Language System (UMLS)\footnote{
    We use the 2016AA release of the UMLS.
} \cite{UMLS} in order
to identify medical term pairs sharing the same relationships.  We followed the standard process for dataset generation outlined in
Section~\ref{ssec:standard_methodology}, with some adjustments for the assumptions in Section~\ref{ssec:assumptions}.

The UMLS Metathesaurus is centered around normalized {\it concepts}, represented by Concept Unique Identifiers (CUIs).
Each concept can be represented in textual form by one or more {\it terms} (e.g., C0009443 $\rightarrow$ ``Common cold'',
``acute rhinitis'').  These terms may be multi-word expressions (MWEs); in fact, many concepts in the UMLS have no unigram
terms.

The Metathesaurus also contains $\langle${\it subject, relation, object}$\rangle$ triples describing binary relationships
between concepts.  These relationships are specified at two levels:
relationship types (RELs), such as {\it broader-than} and {\it qualified-by}, and specific relationships (RELAs) within each
type, e.g., {\it tradename-of} and {\it has-finding-site}.  For this work, we
used the 721 unique REL/RELA
pairings as our source relationships, and treated the $\langle${\it subject, object}$\rangle$ pairs linked within
each of these relationships as candidates for generating analogies.

To enable a word embedding--based evaluation,
we first identified terms that appeared at least 25 times in the 2016 PubMed baseline collection of biomedical abstracts,\footnote{
    We chose 25 as our minimum frequency to ensure that each term appeared often enough to learn reasonable embeddings for its
    component words.  To determine term frequency, we first lowercased and stripped punctuation from both the PubMed corpus and the term
    list extracted from UMLS, then searched the corpus for exact term matches.
} and removed all $\langle${\it subject, object}$\rangle$ pairs involving concepts that did not correspond
to these frequent terms.  
Most relationships in the Metathesaurus are many-to-many (i.e., each subject can be paired with multiple objects
and vice versa), and thus may challenge Single-Target and Informativity assumptions; we therefore next identified relations
that had at least 50 1:1 instances, i.e., a subject and object that are only paired with one another within a specific
relationship.
Since 1:1 instances are not sufficient to guarantee Informativity, we
then manually reviewed the remaining relations to identify those those that we deemed to satisfy Informativity constraints.
For example, the {\it is-a} relationship between {\it tongue muscles} and {\it head muscle} is not specific enough to suggest
that {\it carbon monoxide} should elicit {\it gasotransmitters} as its corresponding answer.  However, for {\it associated-with}, sampled
pairs such as {\it leg injuries : leg} and {\it histamine release : histamine} were sufficiently consistent that we deemed it
Informative.  This gave
us a final set of 25 binary relations, listed in Table~\ref{tbl:kept_relations}.\footnote{Examples of each relation, along
with their mappings to UMLS REL/RELA values, are available online.}

We follow \newcite{Gladkova2016} in generating a balanced dataset, to enable a more robust comparative analysis between relations.
We randomly sampled 50 $\langle${\it subject, object}$\rangle$ pairs from each relation, again restricting to concepts with strings
appearing frequently in PubMed.  For each subject concept that we sampled, we collected all valid object concepts and bundled them
as a single $\langle${\it subject, objects}$\rangle$ pair.  We then exhaustively combined each concept pair with the others in its
relation to create 2,450 analogies, giving us a total dataset size of 61,250 analogies.  Finally, for each concept, we chose a single
frequent term to represent it, giving us both CUI and string representations of each analogy.

\begin{table}[!t]
    \begin{tabular}{|c|p{5.5cm}|c|}
        \hline
        ID&Name&Amb\\
        \hline
        \multicolumn{3}{|l|}{\it Lab/Rx}\\
        \hline
            L1&\LOne    & 1.0\\
            L2&\LTwo    & 1.1\\
            L3&\LThree  & 1.5\\
            L4&\LFour   & 1.3\\
            L5&\LFive   & 1.6\\
            L6&\LSix    & 1.0\\
            L7&\LSeven  & 1.1\\
            L8&\LEight  & 1.3\\
        \hline

        \multicolumn{3}{|l|}{\it Hierarchical}\\
        \hline
            H1&\HOne    & 1.0\\
            H2&\HTwo    &10.4\\
        \hline

        \multicolumn{3}{|l|}{\it Morphological}\\
        \hline
            M1&\MOne    & 1.1\\  
            M2&\MTwo    & 1.0 \\
        \hline

        \multicolumn{3}{|l|}{\it Clinical}\\
        \hline
            C1&\COne    & 2.6\\
            C2&\CTwo    & 1.5\\
            C3&\CThree  & 4.6\\
            C4&\CFour   & 2.0\\
            C5&\CFive   & 1.9\\
            C6&\CSix    & 1.2\\
        \hline

        \multicolumn{3}{|l|}{\it Anatomy}\\
        \hline
            A1&\AOne    & 1.6\\
            A2&\ATwo    & 5.4\\
            A3&\AThree  & 1.4\\
        \hline

        \multicolumn{3}{|l|}{\it Biology}\\
        \hline
            B1&\BOne    & 1.0\\
            B2&\BTwo    & 1.0\\
            B3&\BThree  & 1.1\\ 
            B4&\BFour   & 2.4\\
        \hline
    \end{tabular}
    \caption{List of the relations kept after manual filtering; Amb is the average ambiguity, i.e., the average number of correct answers per analogy.}
    \label{tbl:kept_relations}
\end{table}

\section{Evaluation}

We assess how well biomedical word embeddings can perform on our dataset, and explore modifications to
the standard evaluation methodology to relax the assumptions described in Section~\ref{ssec:assumptions}.
We use the skip-gram embeddings trained by \newcite{Chiu2016b} on the PubMed citation database, one set using
a window size of 2 (PM-2) and another set with window size 30 (PM-30).  All other word2vec hyperparameters were
tuned by Chiu et al.\ on a combination of similarity and relatedness and named entity recognition tasks.

Additionally, we use the hyperparameters they identified (minimum frequency=5, vector dimension=200, negative
samples=10, sample=1e-4, $\alpha$=0.05, window size=2) to train our own embeddings on a subset of the 2016
PubMed Baseline (14.7 million documents, 2.7 billion tokens).  We train word2vec \cite{Mikolov2013a} samples
with the continuous bag-of-words (CBOW) and skip-gram (SGNS) models, trained for 10 iterations, and GloVe
\cite{Pennington2014} samples, trained for 50 iterations.

We performed our evaluation with each of \CosAdd, \PairDist, and \CosMul\ as the scoring function over the
vocabulary.  In contrast to the prior findings of \newcite{Levy2014} on the Google dataset, performance on
BMASS is roughly equivalent among the three methods, often differing by only one or two correct answers.
We therefore only report results with \CosAdd, since it is the most familiar method.

\subsection{Modifications to the standard method}

We consider \CosAdd\ under three settings of the analogies in our dataset.  For a given analogy
{\it a:b::c:?d}, we refer to $\langle a,b\rangle$ as the exemplar pair and $\langle c,d\rangle$
as the query pair; {\it?d} signifies the target answer.

{\bf Single-Answer} puts analogies in {\it a:b::c:d} format, with a single example object $b$ and a
single correct object $d$, by taking the first object listed for each term pair.  This enforces the
Single-Answer assumption.

{\bf Multi-Answer} takes the first object listed for the exemplar term pair, but keeps all valid answers,
i.e. {\it a:b::c:[$d_1$,$d_2$,\dots]}; this is similar to the approach of \newcite{Gladkova2016}.
There are approximately 16k analogies in our dataset with multiple valid answers.

{\bf All-Info} keeps all valid objects for both the exemplar and query pairs.  The exemplar offset is
then calculated over $B = [b_1, b_2, \dots]$ as
\begin{equation*}
    a - B = \frac{1}{|B|}\sum_i a - b_i
\end{equation*}
Though this is superficially similar to \CosAvg\ \cite{Drozd2016}, we average over objects for a specific subject,
as opposed to averaging over all subject-object pairs.

\begin{table}[b]
    \centering
    \small
    \begin{tabular}{|c|ccc|ccc|}
        \hline
        \multirow{2}{*}{Rel}&\multicolumn{3}{c|}{PM-2}&\multicolumn{3}{c|}{CBOW}\\
        &Uni&Uni$_{M}$&MWE&Uni&Uni$_{M}$&MWE\\
        \hline
        L2&0.07&0.10&0.07&0.11&0.14&0.06\\
        L3&0.14&0.19&0.06&0.12&0.16&0.06\\
        L4&0.01&0.00&0.02&0.04&0.05&0.07\\
        \hline
    \end{tabular}
    \caption{MAP performance on the three BMASS relations with $\geq$100 unigram analogies.  Uni is using
             unigram embeddings on unigram data, Uni$_{M}$ is using MWE embeddings on unigram data, and MWE
             is performance with MWE embeddings over the full MWE data.}
    \label{tbl:unigram_comparison}
\end{table}
We report a relaxed accuracy (denoted \RelAcc), in which the guess is correct if it is in the set of correct answers.
(In the Single-Answer case, this reduces to standard accuracy.)  \RelAcc, as with standard accuracy, necessitates
ignoring $a,b,$ or $c$ if they are the top results \cite{Linzen2016}.

In order to capture information about all correct answers, we also report Mean Average Precision (MAP) and Mean
Reciprocal Rank (MRR) over the set of correct answers in the vocabulary, as ranked by Equation~\ref{eq:3CosAdd}.
Since MAP and MRR do not have a cutoff in terms of searching for the correct answer in the
ranked vocabulary, they can be used without the adjustment of ignoring $a,b,$ and $c$; thus, they can give a more
accurate picture of how close the correct terms are to the calculated guesses.

\begin{figure*}[!t]
    \centering
    \includegraphics[width=\textwidth]{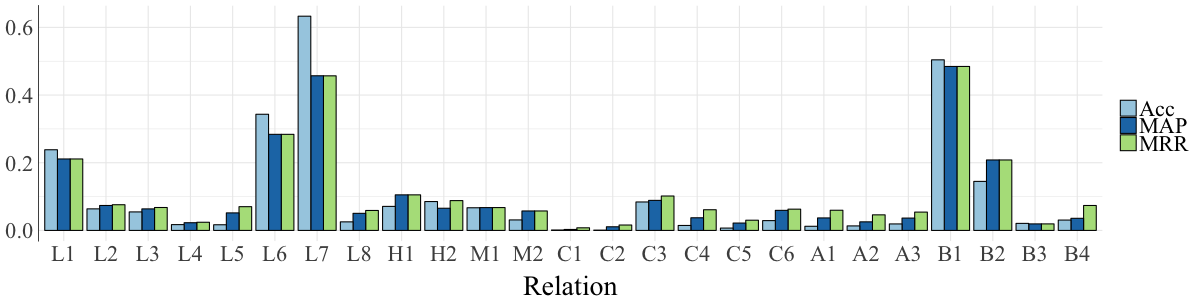}
    \caption{\RelAcc, MAP, and MRR for each relation, using PM-2 embeddings under the Multi-Answer setting.  Note that MAP
        is calculated using the position of all correct answers in the ranked list, while MRR reflects only the position of
        the first correct answer found in the ranked list for each individual query.}
    \label{fig:per_relation_MA}
\end{figure*}
\begin{figure*}[!t]
    \centering
    \includegraphics[width=\textwidth]{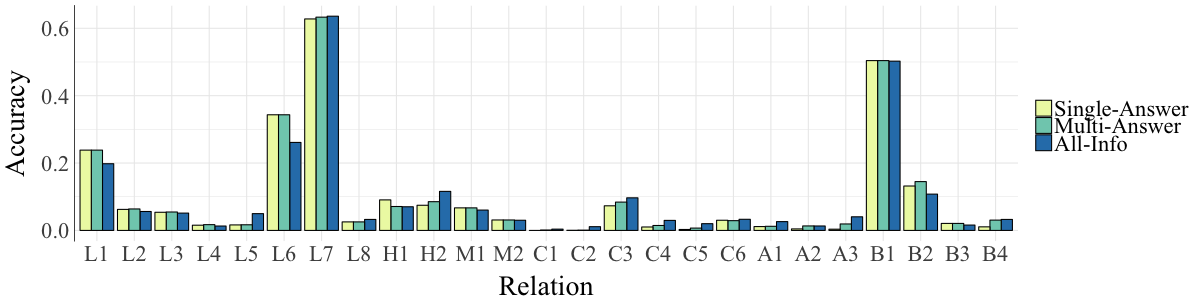}
    \caption{\RelAcc\ per relation for PM-2 on \BMASS, under Single-Answer, Multi-Answer, and All-Info settings.}
    \label{fig:per_relation_acc}
\end{figure*}
\begin{figure*}[!t]
    \centering
    \includegraphics[width=\textwidth]{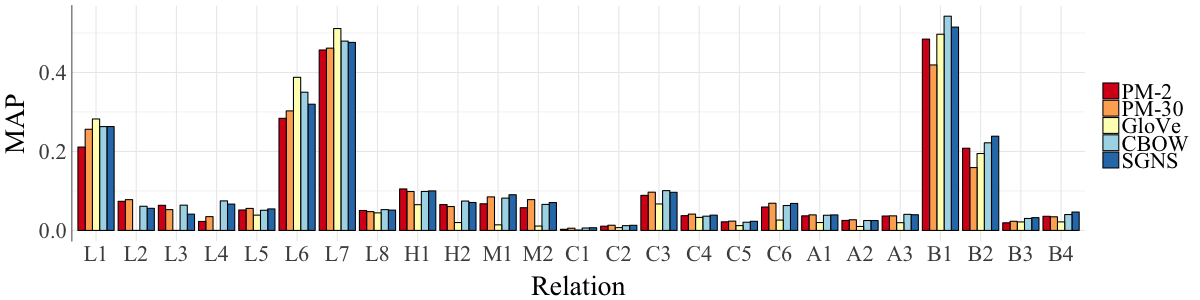}
    \caption{Per-relation MAP for all embeddings under the Multi-Answer setting.}
    \label{fig:per_relation_vectors}
\end{figure*}

\subsection{MWEs and candidate answers}

As noted in Section~\ref{sec:bmass}, the terms in our analogy dataset may be multi-word expressions (MWEs).  We follow the common
baseline approach of representing an MWE as the average of its component words \cite{Mikolov2013b,Chen2013,Wieting2016}.
For phrasal terms containing one or more words that are out of our embedding vocabulary, we only consider the in-vocabulary words: thus,
if ``parathyroid'' is not in the vocabulary, then the embedding of {\it parathyroid hypertensive factor} will be
\begin{equation*}
    \frac{hypertensive + factor}{2}
\end{equation*}

For any individual analogy {\it a:b::c:?d}, the vocabulary of candidate phrases to complete the analogy is derived by calculating
averaged word embeddings for each UMLS term appearing in PubMed abstracts at least 25 times. Terms for which none of the component words
are in vocabulary are discarded.  This yields a candidate set of 229,898 phrases for the PM-2 and PM-30, and 263,316 for our
CBOW, SGNS, and GloVe samples.

Since prior work on analogies has primarily been concerned with unigram data, we also identified a subset of our data for which
we could find single-word string realizations for all concepts in an analogy, using the full vocabulary of our trained embeddings.
Even in the All-Info setting, we could only identify 606 such analogies; Table~\ref{tbl:unigram_comparison} shows MAP results
for PM-2 and CBOW embeddings on the three relations with at least 100 unigram analogies.  The unigram analogies are slightly better
captured than the full MWE data for \relref{\LTwo}{L2} and \relref{\LThree}{L3}; however, lower performance on the unigram subset
in \relref{\LFour}{L4} shows that unigram analogies are not always easier.  We see a small effect from the much larger set of
candidate answers in the unigram case ($>$1m unigrams), as shown by the slightly higher MAP numbers in the Uni$_M$ case.  In
general, it is clear that the difficulty of some of the relations in our dataset is not due solely to using MWEs in the analogies.

\begin{table*}
    \small
    \centering
    \begin{tabular}{|r|ccc|ccc|ccc|}
        \hline
        \multirow{2}{*}{Setting}&\multicolumn{3}{c|}{Single-Answer}&\multicolumn{3}{c|}{Multi-Answer}&\multicolumn{3}{c|}{All-Info}\\
        &\RelAcc&MAP&MRR&\RelAcc&MAP&MRR&\RelAcc&MAP&MRR\\
        \hline
        PM-2&.10 (.16)&.10 (.13)&.10 (.13)&.10 (.16)&.10 (.13)&.11 (.13)&.10 (.15)&.10 (.13)&.11 (.13)\\
        PM-30&.10 (.17)&.10 (.12)&.10 (.12)&.11 (.17)&.10 (.12)&.11 (.12)&.10 (.16)&.10 (.12)&.11 (.12)\\
        GloVe&.11 (.22)&.09 (.15)&.09 (.15)&.11 (.22)&.09 (.16)&.10 (.15)&.10 (.18)&.09 (.16)&.10 (.15)\\
        CBOW&.11 (.18)&.12 (.14)&.12 (.14)&.12 (.18)&.12 (.14)&.12 (.14)&.11 (.17)&.12 (.14)&.13 (.14)\\
        SGNS&.11 (.18)&.11 (.14)&.11 (.14)&.11 (.18)&.11 (.14)&.12 (.13)&.11 (.17)&.12 (.14)&.12 (.13)\\
        \hline
    \end{tabular}
    \caption{Average performance over all relations in the dataset, for each set of embeddings.  Results are reported as ``Mean (Standard deviation)'' for each metric.}
    \label{tbl:summary_statistics}
\end{table*}

\subsection{Metric comparison}

Figure~\ref{fig:per_relation_MA} shows \RelAcc, MAP, and MRR results for each relation in BMASS, using PM-2 embeddings in the
Multi-Answer setting.
Overall, performance varies widely between relations, with all three metrics staying under 0.1\ in the majority of cases;
this mirrors previous findings on other analogy datasets \cite{Levy2014,Gladkova2016,Drozd2016}.

MAP further fleshes out these differences by reporting performance over all correct answers for a given analogy.
This lets us distinguish between relations like \relref{\LSeven}{L7}, where noticeably lower MAP numbers reflect a wider distribution
of the multiple correct answers, and relations like  \relref{\BTwo}{B2} or \relref{\CSix}{C6}, where a low \RelAcc\ reflects many incorrect
answers, but a higher MAP indicates that the correct answers are relatively near the guess.

MRR, on the other hand, more optimistically reports how close we got to finding any correct answer.  Thus, for the
\relref{\CFour}{C4} relation, low \RelAcc\ is belied by a noticeably higher MRR, suggesting that even when we guess wrong,
the correct answer is close.  This contrasts with relations like \relref{\HOne}{H1} or \relref{\CThree}{C3}, where MRR is more
consistent with \RelAcc, indicating that wrong guesses tend to be farther from the truth.  Since most of our analogies (45,178 samples,
or about 74\%) have only a single correct answer, MAP and MRR tend to be highly similar.  However, in high-ambiguity relations
like \relref{\HTwo}{H2}, higher MRR numbers give a better sense of our best case performance.

\subsection{Analogy settings}

To compare across the Single-Answer, Multi-Answer, and All-Info settings, we first look at \RelAcc\ for each relation in BMASS, shown
for PM-2 embeddings in Figure~\ref{fig:per_relation_acc} (the observed patterns are similar with the other embeddings).  Unsurprisingly,
allowing for multiple answers in Multi-Answer and All-Info slightly raises \RelAcc\ in most cases.  What is surprising, however,
is that including more sample exemplar objects in the All-Info setting had widely varying results.  In some cases, such as
\relref{\HTwo}{H2}, \relref{\LFive}{L5}, and \relref{\CFour}{C4}, the additional exemplars gave a noticeable improvement in accuracy.
In others, accuracy actually went down: \relref{\LOne}{L1} and \relref{\LSix}{L6} are the most striking examples, with absolute decreases
of 4\% and 8\% respectively from the Multi-Answer case for PM-2 (the decreases are similar with other embeddings).  Thus, it seems that
multiple examples may help with Informativity in some cases, but confuse it in others.  Taken together with
the improvements seen in \newcite{Drozd2016} from using \CosAvg, this is another indication that any single subject-object
pair may not be sufficiently representative of the target relationship.

\subsection{Embedding methods}

Averaging over all relations, the five embedding settings we tested behaved roughly the same, with our trained embeddings
slightly outperforming the pretrained embeddings of \newcite{Chiu2016b}; summary \RelAcc, MAP, and MRR performances are given in
Table~\ref{tbl:summary_statistics}.  At the level of individual relations, Figure~\ref{fig:per_relation_vectors} shows MAP performance
in the Multi-Answer setting.  The four word2vec samples tend to behave similarly, with some inconsistent variations.  Interestingly,
CBOW outperforms the other embeddings by a large margin in several relations, including \relref{\BOne}{B1} and \relref{\LFour}{L4}.

GloVe varies much more widely across the relations, as reflected in the higher standard deviations in Table~\ref{tbl:summary_statistics}.
While GloVe consistently outperforms word2vec embeddings on \relref{\LSix}{L6} and \relref{\LSeven}{L7}, it significantly underperforms
on the morphological and hierarchical relations, among others.  Most notably, while the word2vec embeddings show minor differences
in performance between the Multi-Answer and All-Info settings, GloVe \RelAcc\ performance falls drastically on \relref{\LOne}{L1} and
\relref{\LSix}{L6}, as shown in Table~\ref{tbl:glove_variations-rerun_3_bl_string}.  However, its MAP and MRR numbers stay similar, suggesting that there
is only a reshuffling of results closest to the guess.

\begin{table}
    \centering
    \small
    \begin{tabular}{|c|ccc|ccc|}
        \hline
        \multirow{2}{*}{Metric}&\multicolumn{3}{c|}{L1}&\multicolumn{3}{c|}{L6}\\        &SA&MA&AI&SA&MA&AI\\        \hline
        \RelAcc&0.49&0.49&0.25&0.62&0.62&0.40\\
        MAP&0.28&0.28&0.28&0.39&0.39&0.39\\
        MRR&0.28&0.28&0.28&0.39&0.39&0.39\\
        \hline
    \end{tabular}
    \caption{\RelAcc, MAP, and MRR performance variation between Single-Answer (SA), Multi-Answer (MA), and All-Info (AI)
             settings for GloVe embeddings on \relref{\LOne}{L1} and \relref{\LSix}{L6}.}
    \label{tbl:glove_variations-rerun_3_bl_string}
\end{table}

\subsection{Error analysis}

Several interesting patterns emerge in reviewing individual {\it a:b::c:?d} predictions.  A number of errors follow directly from our word
averaging approach to MWEs: words that appear in $b$ or $c$ often appear in the predictions, as in {\it gosorelin:\ul{ici} 118630::letrozole:*\ul{ici} 164384}.
Prefix substitutions also occurred, as with {\it mianserin hydrochloride:mianserin::scopolamine \ul{hydrobromide}:*scopolamine \ul{methylbromide}}.

Often, the $b$ term(s) would outweigh $c$, leading to many of the top guesses being variants on $b$.  In one analogy,
{\it sodium acetylsalicyclate:\ul{aspirin}::intravenous immunoglobulins:?immunoglobulin g}, the top guesses were:
{\it *\ul{aspirin} prophylaxis}, {\it *\ul{aspirin}},
{\it *\ul{aspirin} antiplatelet}, and {\it *low-dose \ul{aspirin}}.

In other cases, related to the nearest neighborhood over-reporting observed by \newcite{Linzen2016}, we saw guesses very similar to $c$, regardless of
$a$ or $b$, as with {\it acute inflammations:acutely inflamed::\ul{endoderm}:*embryonic endoderm}; other near guesses included {\it *endoderm cell} and
{\it epiblast}.

Finally, we found several analogies where the incorrect guesses made were highly related to the correct answer, despite not
matching.  One such analogy was {\it oropharyngeal suctioning:substances::thallium scan:?radioisotopes}; the top guess was
{\it *radioactive substances}, and {\it *gallium compounds} was two guesses farther down.  Showing some mixed effect from the
neighborhood of $b$, {\it *performance-enhancing substances} was the next-ranked candidate.
\section{Discussion}

Relaxing the Single-Answer, Same-Relationship, and Informativity assumptions by including multiple correct answers and multiple exemplar
pairs and by reporting MAP and MRR in addition to accuracy paints a more complete picture of how well word embeddings are performing
on analogy completion, but leaves a number of questions unanswered.  While we can more clearly see the relations where we correctly
complete analogies (or come close), and contrast with relations where a vector arithmetic approach completely misses the mark, what
distinguishes these cases remains unclear.  Some more straightforward relationships, such as \relref{\BThree}{B3} and its inverse
\relref{\BFour}{B4}, show surprisingly poor results, while the very broad synonymy of \relref{\HOne}{H1} is captured comparatively
well.  Additionally, in contrast to prior work with morphological relations, \relref{\MOne}{M1} and \relref{\MTwo}{M2} are much more
challenging in the biomedical domain, as we see non-morphological related pairs such as {\it predisposed:disease susceptibility} and
{\it venous lumen:endovenous}, in addition to more normal pairs like {\it sweating:sweaty} and {\it muscular:muscle}.  Further analysis
may provide some insight into specific challenges posed by the relations in our dataset, as well as why performance with \PairDist\ 
and \CosMul\ did not noticeably differ from \CosAdd.

In terms of specific model errors, we did not evaluate the effects of any embedding hyperparameters on performance in BMASS, opting to
use hyperparameter settings tuned for general-purpose use in the biomedical domain.  \newcite{Levy2015b} and \newcite{Chiu2016b}, among
others, show significant impact of embedding hyperparameters on downstream performance.  Exploring different settings may be one way
to get a better sense of exactly what incorrect answers are being highly-ranked, and why those are emerging from the affine organization
of the embedding space.  Additionally, the higher variance in per-relation performance we observed with GloVe embeddings suggests that
there is more to unpack as to what the GloVe model is capturing or failing to capture compared to word2vec approaches.

Finally, while we considered Informativity during the generation of BMASS, and relaxed the Single-Answer assumption in our evaluation,
we have not really addressed the Same-Relationship assumption.  Using multiple exemplar pairs is one attempt to reduce the impact of
confusing extraneous relationships, but in practice this helps some relations and harms others.  \newcite{Drozd2016} tackle this problem
with the LRCos method; however, their findings of mis-applied features and errors due to very slight mis-rankings show that there is
still room for improvement.  One question is whether this problem can be addressed at all with non-parametric models like the vector
offset approaches, to retain the advantages of evaluating directly from the word embedding space, or if a learned model (like LRCos) 
is necessary to separate out the different aspects of a related term pair.

\section{Conclusions}

We identified three key assumptions in the standard methodology for analogy-based evaluations of word embeddings: Single-Answer
(that there is a single correct answer for an analogy), Same-Relationship (that the exemplar and query pairs are related in the
same way), and Informativity (that the exemplar pair is informative with respect to the query pair).  We showed that these assumptions
do not hold in recent benchmark datasets or in biomedical data.  Therefore, to relax these 
assumptions, we modified analogy evaluation to allow for multiple correct answers and multiple exemplar pairs, and reported Mean
Average Precision and Mean Reciprocal Recall over the ranked vocabulary, in addition to accuracy of the highest-ranked choice.

We also presented the BioMedical Analogic Similarity Set (BMASS), a novel analogy completion dataset for the biomedical domain.
In contrast to existing datasets, BMASS was automatically generated from a large-scale database of
$\langle$subject, relation, object$\rangle$ triples in the UMLS Metathesaurus, and represents a number of challenging real-world
relationships.  Similar to prior results, we find wide variation in word embedding performance on this dataset, with accuracies
above 50\% on some relationships such as \rel{\LSeven} and \rel{\BOne}, and numbers below 5\% on others, e.g., \rel{\AOne} and
\rel{\LEight}.

Finally, we are able to address the Single-Answer assumption by modifying the analogy evaluation to accommodate multiple correct
answers, and we consider Informativity in generating our dataset and using multiple example pairs.  However, the Same-Relationship
assumption remains a challenge, as does a more automated approach to either evaluating or relaxing Informativity.  These offer 
promising directions for future work in analogy-based evaluations.

\section*{Acknowledgments}
We would like to thank the CLLT group at Ohio State and the anonymous reviewers for their helpful comments.  
Denis is a pre-doctoral fellow at the National Institutes of Health Clinical Center, Bethesda, MD.

\bibliographystyle{acl_natbib}
\bibliography{references.edit}

\end{document}